\pdfoutput=1

\documentclass[11pt]{article}

\usepackage[final]{acl}

\usepackage{times}
\usepackage{latexsym}

\usepackage[T1]{fontenc}

\usepackage[utf8]{inputenc}

\usepackage{microtype}

\usepackage{graphicx}
\usepackage{float}
\usepackage{verbatim}
\usepackage{listings}
\usepackage{fancyvrb}
\usepackage{pmboxdraw}
\usepackage{subcaption}
\usepackage{booktabs}

\title{Modeling the Graphotactics of Low-Resource Languages Using Sequential GANs}

\author{Isaac Wasserman \\
  Haverford College / Haverford, PA \\
  University of Pennsylvania / Philadelphia, PA \\
  \texttt{isaacrw@seas.upenn.edu} \\}

\begin{document}
\maketitle
\begin{abstract}
  Generative Adversarial Networks (GANs) have been shown to aid in the creation of artificial data in situations where large amounts of real data are difficult to come by. This issue is especially salient in the computational linguistics space, where researchers are often tasked with modeling the complex morphologic and grammatical processes of low-resource languages. This paper will discuss the implementation and testing of a GAN that attempts to model and reproduce the graphotactics of a language using only 100 example strings. These artificial, yet graphotactically compliant, strings are meant to aid in modeling the morphological inflection of low-resource languages.
\end{abstract}

\section{Introduction}
  \subsection{Task}
    In 2019, Anastasopoulos and Neubig made waves with their multilingual morphological inflection model for low resource languages \citep{CMU} that they submitted to the SIGMORPHON 2019 shared task \citep{sigmorphon2019}. All models submitted were pretrained on high resource languages of similar ancestry to the target language, allowing many models to greatly exceed the performance of previous attempts at low-resource morphological inflection. However, what allowed Anastasopoulos and Neubig's model to outperform other submissions was its use of data ``hallucination''.

    To perform this hallucination, they aligned the lemma with its inflected form, extracted the stem, and generated new artificial examples by replacing this stem with randomly generated strings (in the language's alphabet) of equal length.\footnote{The alignment process assumes that the lemma and inflected form share a common substring.} Though this random substitution may seem haphazard, the approach allowed for an additional 10\% accuracy, on average, when tested against versions of the model that only used cross-lingual transfer.

    Surely, a more well informed approach to stem generation would further improve the accuracy of the inflectional model. Given the demonstrated ability of GANs to produce photorealistic, yet completely contrived images, they are potentially ideal for such a task. The experiments detailed in this paper attempt to produce a technique for generating fake word stems that provide more relevant information to the inflectional model from Anastasopoulos and Neubig, 2019 \citep{CMU}, thereby increasing the accuracy of its inflections. By modeling the graphotactics of the target language using a GAN, it should be possible to produce strings that more accurately depict possible character sequences.

  \subsection{Generative Adversarial Networks}
    Generative adversarial networks are a class of unsupervised machine learning architectures, most commonly used for image generation. These networks consist of a generator and a discriminator that are trained simultaneously on a set of data representing a class or domain; this domain could be anything from photos of human faces to time series of hourly temperatures.\footnote{Technically speaking, the generator and discriminator are most often trained one after another on a repeated basis.} The generator is tasked with producing ``fake'' examples that are within this domain without ever seeing any real examples from the training set. Meanwhile, the discriminator is fed a combination of fake examples (from the generator) and real examples and is tasked with classifying them as real or fake. The respective goals of the generator and discriminator constitute a zero-sum game, in which the generator is constantly trying to outsmart the discriminator, while the discriminator hones its ability to distinguish between in-domain and out-of-domain examples.

    Though GANs are most often applied to image data (as in the popular CycleGAN \citep{cyclegan}, StyleGAN \citep{stylegan}, and DiscoGAN \citep{discogan}), the same logic is also applicable to other types of data. For example, in 2017, Esteban et al. developed a pair of recurrent GANs which they applied to medical time series data \citep{rcgan}, and in 2016, Yu et al. developed a GAN architecture made specifically for sequences and language generation, utilizing techniques from reinforcement learning \citep{seqgan}.

    Although GANs are most popularly used for domain transfer, the same basic architecture is also a good candidate for a data augmentation strategy called hallucination in which new (fake) training examples are created based on a small number of existing (real) examples.

    Though theoretically, the addition of these fake examples should not improve the performance of a discriminative model (since they can be no more representative of the true distribution than the examples they are based on), an empirical study by Antoniou et al. observed accuracy improvements on multi-class classification of up to 13\% on benchmark low-resource datasets such as Omniglot \citep{antoniou}. These performance gains are supported by a fairly extensive body of similar evidence-based studies on GAN-based data augmentation in low-resource settings. These studies are, more often than not, concerned with medical imaging, in which segmentation is a more salient issue than classification.

    Shin et al., 2018 \citep{shin} leveraged the popular Pix2Pix \citep{pix2pix} architecture to augment a brain-tumor segmentation dataset. This task was considerably more complex as it involved the hallucination of image pairs. However, training their segmentation model on a combination of real and fake data, they observed performance improvements of up to 16\% over unaugmented data (minimum improvement of ~1\%). They also tested the effects of an entirely synthetic dataset; however, this resulted in a significant loss of performance compared to the baseline. Despite the improvements realized when compared to their baseline segmentor, even their best model was unable to outperform the best-in-class reference model \citep{wang}.

    Sandfort et al., 2019 \citep{sandfort} carried out a similar study using the also popular CycleGAN architecture \citep{cyclegan} to segment anomolous CT scans of kidneys, livers, and spleens. On average, this study showed no appreciable difference between the performance gains afforded by traditional augmentation and GAN-based augmentation. However, when the resulting models were tested on images that were out-of-distribution, they found that while the baseline model scored 0.101, the traditionally augmented and GAN augmented models received scores of 0.535 and 0.747 respectively.\footnote{While the training set was based on CTs with contrast, these out-of-distribution images came from scans performed without contrast.} This increased flexibility is interesting as it suggests that the augmented examples constitute a wider domain than the examples they are based on. From a theoretical standpoint, this would have to be possible if GAN augmented datasets were to outperform others.

    While all of the other related works cited have used GANs to create synthetic image data, Gupta, 2019 \citep{gupta} applies GAN-based augmentation to language data for the purpose of sentiment analysis. Unlike the other applications, labeled language data for sentiment analysis is not scarce. The datasets used each contain between 2000 and 4000 real examples, and the classifier was pre-trained on a dataset of over 1.6 million examples. Real and fake examples were combined in a somewhat nontraditional way; instead of concatenating the real and synthetic datasets, separate classifiers were trained for each and their outputs were combined via bagging. In these experiments, accuracy was only improved by up to 1.2\% when fake examples were added. Though these gains are less impressive than others cited above, these results support the idea that GAN-based augmentation can be applied to domains outside of image data.

\section{Method}
  \subsection{Data Pipeline and Architecture}
    All language data used was sourced from the SIGMORPHON 2019 repository and was initially formatted according the the Unimorph schema. The data for each language was gathered from their respective Wikipedia. Only the ``train-low'' sets were used for training. These datasets have only 100 examples, and each example contains a lemma (base form), an inflected form, and a list of the morphological categories associated with the inflected form.\footnote{Although the ``train-low'' sets do occasionally have 1000 examples, those corresponding to the languages used in this experiment have only 100.} 
  
    To produce training data for the GAN, base forms were aligned with inflected forms using a technique developed by Mans Hulden called Chinese restaurant process alignment \citep{crpalign}. Based on the alignment, a series of likely stems are generated using code from Anastasopoulos and Neubig, 2019 \citep{CMU}; as most of the languages tested use primarily suffixational inflection, the first of these ``likely'' stems was assumed to be the true stem and was included in the language's training set as an example of graphotactic compliance.


    Finally, the stems were padded to reach a length of 10 and encoded as 10 $\times$ $len(\textrm{alphabet})$ one-hot sequences.

    Although proven methods exist for building GANs for sequence data such as SeqGAN \citep{seqgan}, it was decided that these models would likely be too complex and therefore prone to overfitting and vanishing gradients when trained on such a small dataset. Instead, a minimalally complex GAN was created (Fig. \ref{fig:architecture}). The generator took noise input of shape 10 (representing the maximum expected stem length) $\times$ $len(\textrm{alphabet})$ and consisted of two 100 unit LSTMs with $tanh$ activations, separated by a 20\% dropout layer, and concluded with a single $len(\textrm{alphabet})$ unit dense layer with softmax activation, producing an output of equal shape to the input. The discriminator receives this output concatenated with one real example from the training set. Due to signs of mode collapse in early tests with two layers, the discriminator was given just one 100 unit LSTM. The discriminator terminates with a sigmoid activated single unit dense layer, that outputs a predicted probability that each example is real. The models use Wasserstein loss \citep{wgan} to reduce the probability of mode collapse.



  \subsection{Testing}
    The resultant models were each used to produce 10,000 artificial training examples. This process involved a modified version of the augmentation script from Anastasopoulos and Neubig, 2019 \citep{CMU}.
    
    The strings produced by the resultant models were, unfortunately, not always of the correct shape. Some had zeroes (representing null characters) in the middle, others filled the entire 10 available characters (which most real stems do not). To combat this, model output strings were ``cleaned''. Strings were cut off at the first zero from the left (all languages were left-to-right), if a character was repeated more than twice consecutively, it was collapsed to just two, and if the length of the stem was longer than the real stem it was replacing, it was shortened to a matching length. 

    The inflectional model from Anastasopoulos and Neubig, 2019 \citep{CMU} was used as a benchmark. For each language, three sets of ``hallucinated'' data were generated. The first used the random character selection method from Anastasopoulos and Neubig, 2019 \citep{CMU}, the second was the GAN produced set, and the third was produced by a simple trigram model trained on same 100 example as the GAN.


    \begin{figure*}[t]
      \center
      \begin{tabular}{lrrrr}
        \toprule
        Language & Unaugmented & Random & Trigram & GAN \\
        \midrule
        Azeri & 3.13 & \textbf{1.76} & 2.02 & 5.42 \\
        Bengali & 2.66 & \textbf{0.81} & 0.95 & 4.59 \\
        Crimean Tatar & 2.34 & 0.54 & \textbf{0.50} & 3.18 \\
        Karelian & 1.74 & 1.94 & \textbf{1.46} & 4.10 \\
        Kashubian & 1.44 & \textbf{1.22} & \textbf{1.22} & 2.78 \\
        Livonian & 3.91 & \textbf{2.23} & 2.55 & 4.78 \\
        Maltese & 2.31 & \textbf{1.51} & 1.91 & 5.29 \\
        Middle High German & 0.80 & \textbf{0.46} & 0.60 & 1.92 \\
        North Frisian & \textbf{3.12} & 3.21 & 3.54 & 7.88 \\
        Occitan & 2.19 & \textbf{0.53} & 0.57 & 5.78 \\
        Old Church Slavonic & 2.34 & \textbf{1.09} & 1.94 & 3.00 \\
        Pashto & 2.53 & \textbf{1.25} & 1.34 & 4.79\\
        Tatar & 2.19 & \textbf{0.50} & 0.62 & 5.52\\
        \midrule
        Average & 2.36 & \textbf{1.31} & 1.48 & 4.54\\
        \bottomrule
      \end{tabular}
      \caption{Average Levenshtein distance (minimum number of elementary string operations to go from prediction to label) of resultant inflectional models using various data augmentation methods}
      \label{fig:lev}
    \end{figure*}

    \section{Results}
    \subsection{GAN Training}
      Models generally fell into two camps when training. With some, the generator loss quickly rose to 1 (the maximum) while the discriminator loss simultaneously fell to -1 (the minimum). With the others, generator loss fluctuated rapidly between 0 and 1, while the discriminator loss fluctuated rapidly between -1 and 0.
      

      In the first case, it seems that the model is learning for the first few epochs but quickly exhausts its potential. This is evident in the sample strings generated as well. As an example, for the first 100 or so epochs, the Kashubian model generated strings that were mostly a single character: ``şşşşşşşşşş''. But after about 300 epochs, string began to resemble stems with a reasonable number of zeroes (null characters) at the end: ``dümç000000''.

      In the second case, the model is able to return to these more realistically structured strings occasionally (denoted by the fluctuations), but it is not able to produce these strings with any consistency.

    \subsection{Performance}
      
      Neither the GAN nor the trigram model were able to consistently produce data that was more informative than the random strings (Fig. \ref{fig:lev}). For all languages except North Frisian and Occitan, training on the randomly augmented set resulted in the highest accuracy. The models trained on the trigram augmented set often performed similarly to those trained on the random set, having identical accuracy in Occitan and slighly higher accuracy in North Frisian.

      No model trained on GAN augmented data achieved higher than 18\% accuracy, and in 10 of 13 cases, these models were less than 10\% accurate. In all cases, training on the 100 example ``train-low'' set alone resulted in a more accurate model than training on both the ``train-low'' and GAN augmented sets; on average, adding GAN augmented data resulted in a model that was only one fifth as accurate as models trained on unuagmented data.

      On average Levenshtein distance calculations follow a similar pattern to that of model accuracy; however, this metric occasionally favors the models trained on trigram augmented data (Fig. \ref{fig:lev}). Additionally, in the case of North Frisian, a particularly difficult language to inflect (based on the accuracy results), the model trained on just the 100 original examples had a slightly lower average Levenshtein distance.

  \section{Discussion}
    \subsection{Analysis}
      GAN based augmentation, using the techniques described, is not a good option for creating artificial training data for the inflection model. It seems as though 100 examples is simply insufficient for the GAN to learn properly. Though there was initially concern that GAN based augmentation could only produce redundant examples, providing no additional information, the fact that it results in worse inflection performance than using unaugmented data means that the GAN is providing novel, yet misleading, information.
      

      Anastasopoulos and Neubig's method, random augmentation, remains the best method for creating informative yet nondeceptive artificial examples \citep{CMU}. The randomness indicates to the inflection model that the stem varies freely and has no bearing on the inflection. However, this is, of course, not always the case. Though many languages have ``regular'' inflectional paradigms (adding ``-ed'' for past tense in English), there are also many irregular forms, influenced by the stem (``run'' [+past] = ``ran''). In these cases, the stem must be taken into consideration to predict the correct inflectional morpheme. This means that an ideal informed (not entirely random) model should be able to produce more accurate and informative data than the random augmentor.

      

    \subsection{Conclusion}
      The emergence of new machine learning technologies are an opportunity to revisit attempts at low-resource language preservation and support. As language speakers and language learners, we know that a dataset of thousands of examples is not a necessary prerequisite for learning the inflectional paradigms of natural language. There is no doubt that small datasets like those used in the experiments above will one day be sufficient for inflection automation with human-like accuracy. 

\bibliography{anthology,custom}
\bibliographystyle{acl_natbib}
\end{document}